\documentclass{article} % For LaTeX2e
\usepackage{iclr2023_conference,times}

\usepackage{amsmath,amsfonts,bm}

\def\eqref#1{equation~\ref{#1}}
\def\1{\bm{1}}

\DeclareMathAlphabet{\mathsfit}{\encodingdefault}{\sfdefault}{m}{sl}
\SetMathAlphabet{\mathsfit}{bold}{\encodingdefault}{\sfdefault}{bx}{n}

\usepackage{hyperref}
\usepackage{url}
\usepackage{adjustbox}
\usepackage{multirow}

\usepackage[utf8]{inputenc} % allow utf-8 input
\usepackage{hyperref}       % hyperlinks
\usepackage{url}            % simple URL typesetting
\usepackage{booktabs}       % professional-quality tables
\usepackage{amsfonts}       % blackboard math symbols
\usepackage{nicefrac}       % compact symbols for 1/2, etc.
\usepackage{microtype}      % microtypography
\usepackage{xcolor}         % colors
\usepackage{amsmath}
\usepackage{graphicx}
\usepackage{subfigure}
\usepackage{amssymb}
\usepackage{mathtools}
\usepackage{mathrsfs}
\usepackage{hyperref}
\usepackage{float}
\usepackage{multirow}
\usepackage{wrapfig}
\usepackage[textsize=tiny,textwidth=25mm]{todonotes}

\title{Predicting Out-of-Distribution Error with Confidence Optimal Transport}

\iclrfinalcopy
\author{Yuzhe Lu$^1$, Zhenlin Wang$^{1}$, Runtian Zhai$^{1}$ \\
\textbf{Soheil Kolouri$^2$, Joseph Campbell$^1$, Katia Sycara$^{1}$} \\
$^1$Carnegie Mellon University \ \ $^2$Vanderbilt University \\ 
{\small \texttt{\{yuzhelu, zhenlinw, rzhai, jcampbell, sycara\}@cs.cmu.edu} } \\ 
{\small \texttt{soheil.kolouri@vanderbilt.edu}}
}

\begin{document}

\maketitle

\begin{abstract}
Out-of-distribution (OOD) data poses serious challenges in deployed machine learning models as even subtle changes could incur significant performance drops. Being able to estimate a model's performance on test data is important in practice as it indicates when to trust to model's decisions. 
% \todo{Talk about why OOD performance estimation is important in practice (it tells us whether we can apply a trained model to a certain task, or serves as a warning in a continual learning setting if the performance drops too much).} 
We present a simple yet effective method to predict a model's performance on an unknown distribution without any addition annotation. Our approach is rooted in the Optimal Transport theory, viewing test samples' output softmax scores from deep neural networks as empirical samples from an unknown distribution. We show that our method, Confidence Optimal Transport (COT), provides robust estimates of a model's performance on a target domain. Despite its simplicity, our method achieves state-of-the-art results on three benchmark datasets and outperforms existing methods by a large margin. 
\end{abstract}

\section{Introduction}
While most machine learning models are built with an i.i.d. assumption, they often have to deal with out-of-distribution samples when deployed in the real world. This poses serious problems on the trustworthiness of ML models as modern architectures often give overly confident signals even in the face of distribution shifts that significantly hurt the model's performance. Thus, a lot of works have focused on OOD detection, whose goal is to tell if a test sample is from a different distribution than the training data. To name a few, \cite{hendrycks2016baseline} proposed to use a model's confidence based on softmax scores; \cite{liang2017enhancing} found using temperature scaling \cite{guo2017calibration} and small perturbations to inputs further improves OOD detection using softmax scores; more recently, \cite{liu2020energy} showed superior performance by using energy-based scores instead of softmax scores. In a relevant but slightly different formulation, \cite{platanios2016estimating, platanios2017estimating, jiang2021assessing, garg2022leveraging} considers the task of directly predicting a model's generalization performance on OOD data. As annotating data from the target domain would be both costly and inefficient, especially when the distribution is non-stationary, it would be ideal to have a good estimator of a model's performance on the target domain in an annotation-free manner. However, previous works have established that obtaining a consistent estimator is impossible without further assumptions on the nature of the shift \cite{david2010impossibility, lipton2018detecting}.

% \todo{Also talk about OOD detection, which was studied earlier than performance estimation.} 

In this work, we aim to estimate the model performance under covariate shift, such as different appearances of the same object, differences in image quality caused by camera degradation, etc., 
% \todo{It would be great to put some examples here (e.g. similar to Figure 1 in the OCL-PDS paper)}
while assuming the label distribution of the target domain does not change. With the covariate shift assumption, 
% \todo{Emphasize more about the covariate shift assumption.} 
the task of predicting model performance with only labeled source data and unlabeled target data becomes achievable. To deal with this task, numerous methods have been proposed. Among them, many require some labeled samples from the target domain. For example, \cite{guillory2021predicting} and \cite{deng2021labels} train regression models based on model-based distribution distances between the source and target domain to predict test performance. However, these methods are suboptimal as they assume prior knowledge about the nature of the shift and obtaining labels from the target domain is very expensive in general. Meanwhile, more recent methods have focused on predicting target domain performance without requiring additional labels. \cite{baekagreement} and \cite{jiang2021assessing} leverage the agreement rate of independently trained neural networks to estimate the accuracy. \cite{garg2022leveraging} adopts a thresholding strategy based on scores of the softmax outputs. These methods have immediate practical value as people can have a sense of their model's performance on the target domain without having to label any data beforehand. Following these works, we assume access to labeled source data and only unlabeled target domain data.

Given this setting, we propose Confidence Optimal Transport (COT) to provide robust estimation of a model's error on the target domain via the Earth Mover's Distance (EMD). While other methods also explored using distribution distance such as Maximum Mean Discrepancy and Fréchet distance for the task \cite{guillory2021predicting}, they mainly use first and second order summary statistics of source and target data features from a model's penultimate layer. Moreover, the output distances are not calibrated, meaning labeled data from the target domain is required to learn an additional regression model, and often underperform on shifts the regression model was not trained on \cite{garg2022leveraging}. In contrast, our method considers the cost to transform samples from the source to the target domain, leveraging information from every sample instead of only the mean and covariance. By representing each sample through its softmax vector, we empirically show that the resulting EMD between the source and target data after de-biasing serves as a highly robust estimator of a model's performance on the target domain and outperforms previous methods on three benchmark OOD datasets.

\section{Method}

\subsection{Optimal Transport and Wasserstein Metric}
In recent years, optimal transport theory has found numerous applications in the field of machine learning. Optimal transport aims to move one distribution of mass to another as efficiently as possible under a given cost function. Wasserstein distances are rooted in the optimal transportation problem, providing a robust mathematical framework for comparing probability distributions that respect the underlying space geometry. Specifically, the p-Wasserstein distance between two probability measures $\mu$ and $\nu$ is defined as:

\begin{equation}
    W_{p}(\mu, \nu) = \left( \inf_{\pi \in \Pi(\mu, \nu)} \int \| x - y\|^{p}d\pi(x, y) \right) ^{1/p}
\end{equation}

where $\Pi(\mu, \nu)$ is the set of all possible couplings. However, in most applications, we have to work with empirical measures. Let $X_1, ..., X_n \sim P$, and $Y_1, ..., Y_m \sim Q$ be i.i.d. samples, and define the empirical measures $P_n = 
\frac{1}{n}\sum_{i = 1}^{n}\delta_{X_i}$ and $Q_n = \frac{1}{m}\sum_{i = 1}^{m}\delta_{Y_i}$. Then the empirical estimator of the p-Wasserstein distance is:

\begin{equation}\label{eq:ewd}
    W_{p}(P_n, Q_m) = \left(\inf_{\pi \in \Pi(P_n, Q_m)} \sum_{i, j = 1}^{m, n} \|X_i - Y_j\|^p\pi_{ij}\right) ^{1/p}
\end{equation}

which can be solved as a linear assignment problem. In particular, when $p=1$, the metric (1-Wasserstein distance) is commonly known as the Earth Mover's Distance (EMD), and the resulting cost metric between samples is Manhattan distance. In the following section, we will demonstrate how we leverage EMD to tackle the problem of estimating a model's performance on test data.

\subsection{Confidence Optimal Transport}
In this section, we introduce our method, Confidence Optimal Transport (COT), that leverages the optimal transport framework to predict the performance of a trained classifier on the target domain consisting of only unlabeled data $X^{t} = \{x^{t}_{1:m}\}$.  In its essence, COT uses the empirical estimator of the Earth Mover's Distance between labels from the source domain and softmax outputs of samples from the target domain to provide highly accurate estimates.

For COT to provide unbiased estimates, the classifier needs to be first calibrated on a validation set from the source domain. Let $C$ be our classifier of interest. Given validation set $\{X^{s}, Y^{s}\}$ from the source, we say $C$ is perfectly-calibrated if $C(X^{s}) = (\hat{Y^{s}}, \hat{P^{s}})$ satisfies the following equation:

\begin{equation}\label{eq:mc}
    \mathbb{P} (\hat{Y^{s}} = Y^{s} | \hat{P} = p) = p, \quad \forall p \in [0, 1]
\end{equation}

where $\hat{Y^{s}}$ denotes class predictions, $\hat{P^{s}}$ denotes associated confidences (probability of correctness), and probability $p$ is over the joint distribution. In practice, Eq. \ref{eq:mc} implies that when a model is perfectly-calibrated, the averaged confidence should be equal to its accuracy. We used Temperature Scaling proposed in \cite{guo2017calibration} to perform model calibration.

 Now suppose classifier $C$ is perfectly-calibrated on the source distribution, and let $S^{t} = \{softmax(C(x_{i:m}^{t}))\}$ be the softmax outputs of our classifier $C$ on the target dataset $X^{t} = \{x^{t}_{1:m}\}$. The estimated error on the target domain $X^{t}$ via COT is defined as:

\begin{equation}\label{eq:cot}
        \widehat{TestError} = \frac{1}{2}EMD(S^{t}, Y^{s}) = \frac{1}{2}\min_{\pi \in \Pi(S^{t}, Y^{s})} \sum_{i, j = 1}^{m, n} \|s_i - y_j\|\pi_{ij}
\end{equation}

where $Y^{s} = \{y^{s}_{1:n}\}$ are simply labels of the validation data from the source domain in one-hot format. Thus, while COT computes a distance metric between the source and target domain, it doesn't require saving any samples from the source domain. In fact, instead of using $Y^{s}$, we can even avoid saving $Y^{s}$ by generating $\tilde{Y}^{s}$ on the fly given knowledge of the label distribution. 
% \todo{A conference submission requires more theoretical intuition behind why COT can be much better than previous methods. Probably not that important for a workshop submission.} 

% Intuitively, the divide-by-two term in Equation \ref{eq:cot} adjusts for the fact that when we compute the $L_1$ norm between a softmax output and a corresponding label in one-hot format, we count the error twice. For example, a calibrated model is $80\%$ sure an image belongs to a class; when the softmax output is matched to the right label, the $L_1$ norm, or Manhattan distance, will be $0.4$ instead of $0.2$, which should be the right estimate.

% \joe{This is an unbiased estimator?}

% \joe{How many samples from the test distribution do you need to have a calibrated test error estimate? How many labels from the source distribution do you need to keep around?}

% \joe{How is the computational complexity of this method compared to other methods? In the conclusion you say it's cubic, is this worse than other methods? How large does n need to be?}

\begin{table*}
    \caption{We compute coefficients of determination ($R^2$) and rank correlations ($\rho$) to measure the linear correlation between a method's output quantity and the actual test error (higher the better). COT achieves superior performance than all existing methods across different models and datasets.}
    \vspace{10pt}
      \centering
      \begin{adjustbox}{width=\textwidth}
        \begin{tabular}{c c c c c c c c c c  c c c c}
        \toprule
        \multirow{2}{*}{Dataset} & \multirow{2}{*}{Network} & \multicolumn{2}{c}{AC} & \multicolumn{2}{c}{Entropy} & \multicolumn{2}{c}{GDE} & \multicolumn{2}{c}{ATC} & \multicolumn{2}{c}{ProjNorm} & \multicolumn{2}{c}{COT} \\
        \cmidrule{3-14}
        $\quad$ & $\quad$ & $R^2$ & $\rho$ & $R^2$ & $\rho$ & $R^2$ & $\rho$ & $R^2$ & $\rho$ & $R^2$ & $\rho$ & $R^2$ & $\rho$ \\
        \midrule
         \multirow{4}{*}{CIFAR10} & ResNet18 & 0.825 & 0.980 & 0.862 & 0.982 & 0.842 & 0.981 & 0.875 & 0.987 & 0.947 & 0.988 & \textbf{0.996} & \textbf{0.998}\\
         $\quad$ & ResNet50 & 0.950 & 0.995 & 0.949 & 0.995 & 0.959 & 0.995 & 0.885 & 0.989 & 0.936 & 0.989 & \textbf{0.993} & \textbf{0.996} \\
         $\quad$ & VGG11 & 0.710 & 0.938 & 0.762 & 0.958 & 0.723 & 0.948 & 0.548 & 0.851 & 0.756 & 0.949 & \textbf{0.994} & \textbf{0.993} \\
         \cmidrule{2-14}
         $\quad$ & Average & 0.828 & 0.971 & 0.858 & 0.978 & 0.841 & 0.975 & 0.769 & 0.942 & 0.880 & 0.975 & \textbf{0.994} & \textbf{0.996} \\
        \midrule
         \multirow{4}{*}{CIFAR100} & ResNet18 & 0.943 & 0.987 & 0.932 & 0.984 & 0.950 & 0.988 & 0.927 & 0.985 & 0.969 & 0.974 & \textbf{0.995} & \textbf{0.997} \\
         $\quad$ & ResNet50 & 0.957 & 0.987 & 0.948 & 0.984 & 0.962 & 0.989 & 0.955 & 0.991 & 0.982 & 0.991 & \textbf{0.992} & \textbf{0.996} \\
         $\quad$ & VGG11 & 0.794 & 0.959 & 0.821 & 0.973 & 0.870 & 0.978 & 0.736 & 0.975 & 0.653 & 0.849 & \textbf{0.996} & \textbf{0.997} \\
         \cmidrule{2-14}
         $\quad$ & Average & 0.898 & 0.978 & 0.900 & 0.980 & 0.927 & 0.985 & 0.873 & 0.984 & 0.868 & 0.938 & \textbf{0.994} & \textbf{0.997} \\
        \midrule
         $\quad$ & ResNet18 & 0.755 & 0.923 & 0.664 & 0.892 & 0.777 & 0.913 & 0.751 & 0.935 & 0.973 & 0.990 & \textbf{0.995} & \textbf{0.998} \\
          {Tiny} & ResNet50 & 0.844 & 0.975 & 0.776 & 0.954 & 0.892 & 0.981 & 0.800 & 0.962 & 0.983 & 0.990 & \textbf{0.994} & \textbf{0.997} \\
          {ImageNet} & VGG11 & 0.563 & 0.736 & 0.493 & 0.684 & 0.587 & 0.748 & 0.746 & 0.837 & 0.671 & 0.825 & \textbf{0.987} & \textbf{0.992} \\
         \cmidrule{2-14}
         $\quad$ & Average & 0.727 & 0.878 & 0.644 & 0.843 & 0.752 & 0.881 & 0.766 & 0.911 & 0.876 & 0.935 & \textbf{0.992} & \textbf{0.996} \\
        \bottomrule
        \end{tabular}
    \end{adjustbox}
    \label{tab:corr-results}
\end{table*}

\begin{figure}[]
    \centering
    \includegraphics[width=\linewidth]{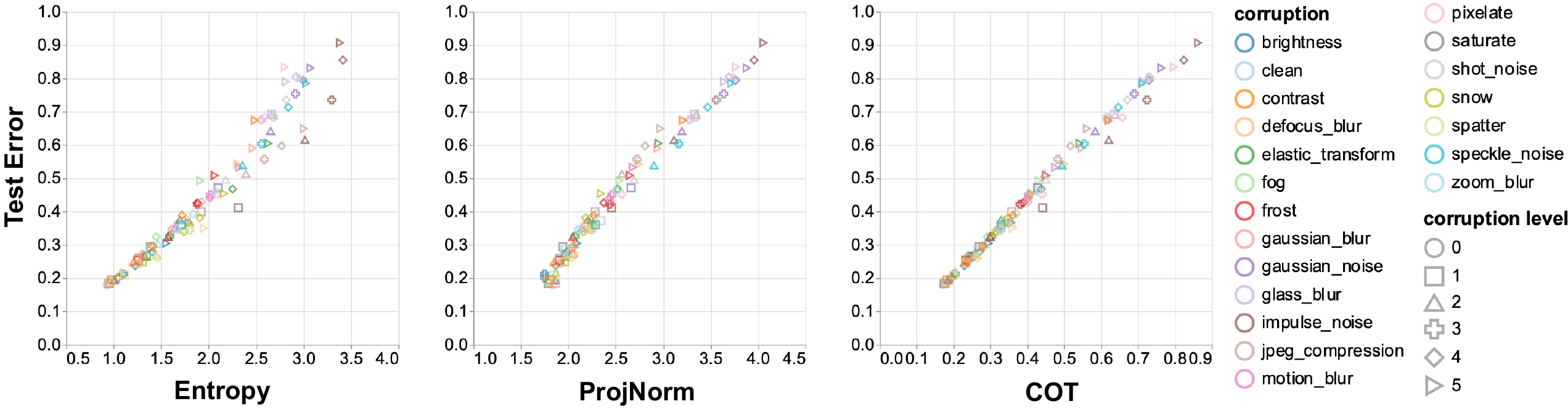}
    \caption{We plot the output metric value against the true test error on CIFAR-100C using ResNet50. Compared to Entropy and ProjNorm, COT shows stronger linear correlation.}
    \label{fig:correlation}
\end{figure}

Next, we will provide some intuitions on Eq. \ref{eq:cot} as well as why we require the classifier to be calibrated. Image a model with $80\%$ accuracy on the source validation set. When it is calibrated, the model will output softmax scores such as [0.8, 0.1, 0.1], [0.1, 0.8, 0.1] on samples from the source test set. When calculating the coupling $\pi$ in \ref{eq:ewd} to move one distribution to another, softmax output [0.8, 0.1, 0.1] will be matched to label [1, 0, 0] and [0.1, 0.8, 0.1] to [0, 1, 0]. Thus, when using Manhattan distance (which is ground cost used by EMD), the cost for a single pair will be $(1 - 0.8) + (0.2 - 0) = 0.4$, which is double the error. Thus, we use EMD (1-Wasserstein distance) and divide it by 2 in Eq. \ref{eq:cot}. At this point, it should be clear that model calibration is a necessity: COT will underestimate the error when the model is overconfident and overestimate when it is underconfident. As models usually become less confident on out-of-distribution samples \cite{hendrycks2016baseline}, we expect COT to output larger error estimates on OOD data. Meanwhile, it is natural to expect COT to outperform simple baselines such as Averaged Confidence, which only utilizes the average confidence instead of the whole softmax output like COT does. 

In terms of computational complexity, since COT requires calculating the EMD, its computation time is $\mathcal{O}(n^3)$. While such complexity seems rather unscalable at the first glance, we will show in Sec. \ref{para:sc} that COT only requires 2000 samples to provide state-of-the-art performance. Thus, when having a large amount of data, we could break it into sufficiently-large subsets and simply apply COT multiple times to obtain a good estimate. 

% adopt a mini-batch strategy to apply COT on a large amount of data from the target domain. \todo{Need to be cautious about this statement, because the experiments show that COT does not work well with insufficient samples.}

\section{Experiments}
We evaluated COT on three OOD benchmark datasets and compared it against several existing methods. COT shows state-of-the-art performance on all evaluation metrics

\paragraph{Dataset} For image classification tasks, we utilized three synthetic-shift datasets, CIFAR10, CIFAR100, Tiny ImageNet as in-distribution data and their corrupted versions, CIFAR10-C, CIFAR100-C, and Tiny ImageNet-C \cite{hendrycks2018benchmarking}, as out-of-distribution datasets. Each standard dataset contains several types of common corruptions in their corrupted counterparts such as change of brightness, existence of fog, occurrence of defocus blur, etc., where each corruption type has 5 levels of severity.

\begin{table}[]
    \centering
    \caption{We use Mean Absolute Error (MAE) to measure the difference between predicted test error and the true test error across different architectures and benchmark datasets. COT, together with AC, GDE, and ATC could provide direct estimates, in contrast to Entropy and ProjNorm, which only demonstrate linear correlation.
     % \todo[inline]{Briefly explain what the MAE here is in the caption. Emphasize that COT can predict the exact accuracy, not just strong linear correlation.}
    }
        \vspace{10pt}
        \begin{tabular}{c c c c c c c c}
        \toprule
        {Dataset} & {Network} & {AC} & {GDE} & {ATC} & {COT} \\
        \cmidrule{1-6}
        \multirow{4}{*}{CIFAR10} & ResNet18 & 10.6 & 13.1 & 3.68 & \textbf{1.9} \\
        $\quad$ & ResNet50 & 6.5 & 9.65 & 3.17 & \textbf{1.50} \\
        $\quad$ & VGG 11 & 10.4 & 22.5 & 6.94 & \textbf{1.94} \\
        \cmidrule{2-6}
        $\quad$ & Average & 9.2 & 15.1 & 4.6 & \textbf{1.8} \\
        \cmidrule{1-6}
        \multirow{4}{*}{CIFAR100} & ResNet18 & 11.3 & 14.9 & 3.6 & \textbf{2.9} \\
        $\quad$ & ResNet50 & 11.2 & 19.2 & 3.5 & \textbf{3.3} \\
        $\quad$ & VGG 11 & 11.7 & 31.2 & 6.2 & \textbf{2.2} \\
        \cmidrule{2-6}
        $\quad$ & Average & 11.4 & 21.8 & 4.5 & \textbf{2.8} \\
        \cmidrule{1-6}
        \multirow{4}{*}{TinyImageNet} & ResNet18 & 12.45 & 27.5 & 4.48 & \textbf{3.43} \\
        $\quad$ & ResNet50 & 13.7 & 29.8 & 3.49 & \textbf{3.24} \\
        $\quad$ & VGG 11 & 12.68 & 43.53 & 7.41 & \textbf{2.02}  \\
        \cmidrule{2-6}
        $\quad$ & Average & 12.95 & 33.60 & 5.12 & \textbf{2.90} \\
        \bottomrule
        \end{tabular} 
        \label{tab:mae-results}
\end{table}

\begin{figure}[]
    \centering
    \includegraphics[width=0.7\linewidth]{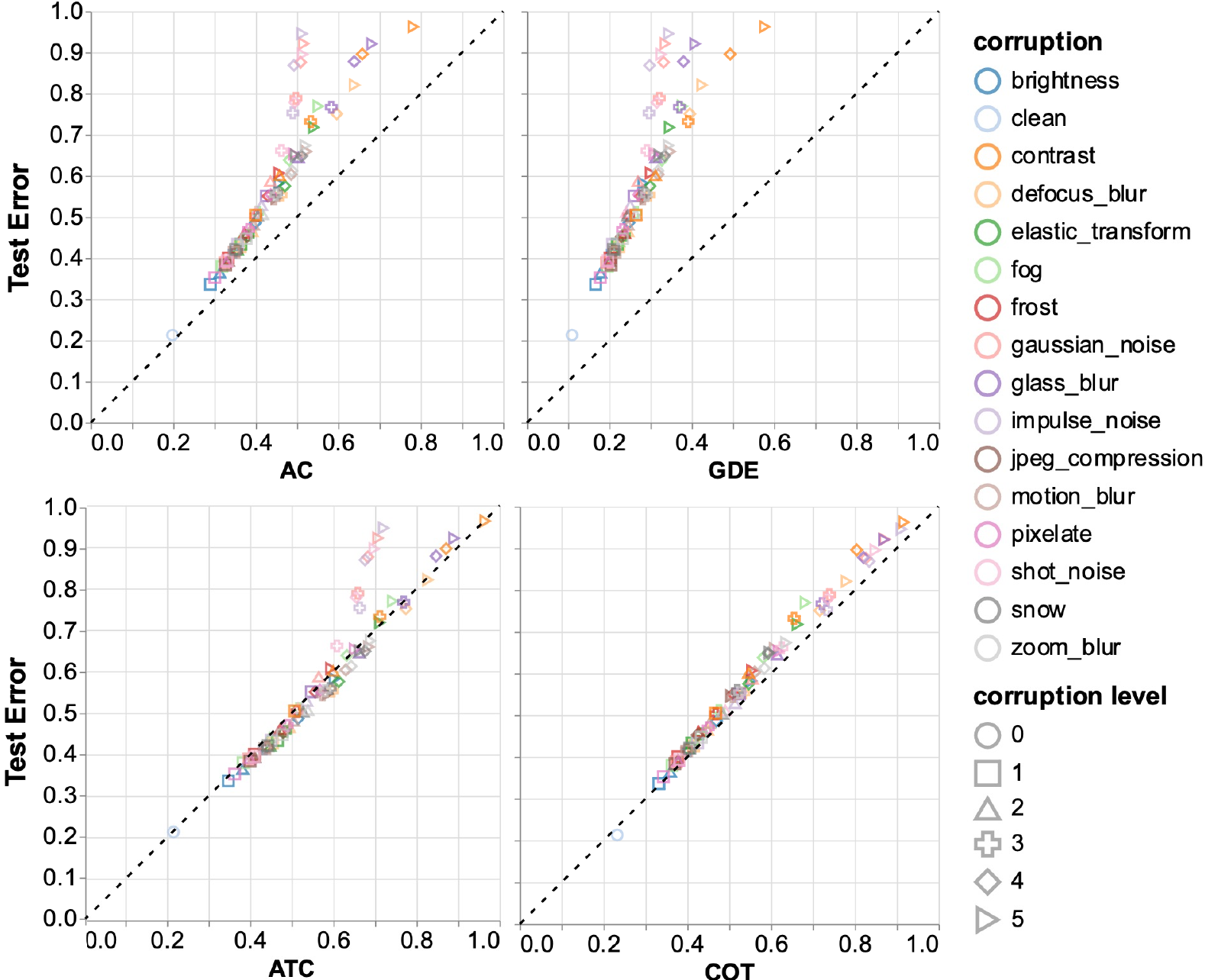}
    \caption{We compare COT's prediction accuracy compared to several other methods. The scattered points are estimates against true error. Since output metric value should be a direct estimate of the test error, we added a dashed black line denoting $y=x$. We can see that COT provides consistent estimates, especially when the shifts are large.}
    \label{fig:mae}
\end{figure}

\paragraph{Architectures} We ran experiments using three different architectures ResNet18, ResNet50 \cite{he2016deep}, and VGG11\cite{simonyan2014very}, for all three datasets. We reserved $10,000$ images as validation set from each dataset. For each architecture, we load the weights pretrained on ImageNet1k \cite{russakovsky2015imagenet}, and fine-tune on the target dataset. Since the datasets used in our experiments have low resolution ($32 \times 32$ for CIFAR, and $64 \times 64$ for Tiny ImageNet), we upsampled the images to $224 \times 224$. After fine-tuning, we calibrated the model on the validation set using Temperature Scaling \cite{guo2017calibration}.

\paragraph{Methods} We consider five existing methods to compare against COT: Averaged Confidence (AC) \cite{hendrycks2016baseline}, Entropy \cite{wang2020tent}, Generalized Disagreement Equality (GDE) \cite{jiang2021assessing}, Averaged Threshold Confidence (ATC) \cite{garg2022leveraging}, and Projection Norm (ProjNorm) \cite{yu2022predicting}. AC is a simple baseline that uses the a model's average confidence to predict performance. Entropy shows that there is strong linear correlation between the entropy of a model's softmax outputs and its performance on OOD data. GDE shows that we can evaluate a model's generalization performance through its disagreement rate with another model independently trained with a different seed. ATC adopts a thresholding strategy by first learning a threshold of the score function of a model's softmax outputs on the validation set and then using the percentage of samples with scores below the threshold as the estimated error rate. Finally, ProjNorm shows that a model's performance on OOD data is linearly correlated to its difference in parameter space from a reference model that's trained with pseudo labels on the target OOD data. It's worth noting that Entropy and ProjNorm only demonstrates linear correlation between the output quantity and the test error, while AC, GDE, ATC, and our method COT not only show linear correlation but also give a direct estimate the test error.

\paragraph{Evaluation Metrics} 
% \todo{Say that these metrics follow prior work.} 
We utilized three evaluation metrics following prior works \cite{yu2022predicting, garg2022leveraging}. The first two, coefficient of determination ($R^2$) and Spearman's rank correlation coefficient ($\rho$), measure the linear correlation between a method's output quantity and the actual test error. The third one,  Mean Absolute Error (MAE), measures how good a method's estimate of the test error is compared to the true error. We adopted $R^2$ and $\rho$ as evaluation metrics as methods such as Entropy and ProjNorm do not give a direct estimate of the error; rather, they output a quantity that's linearly correlated with the error. These methods are less ideal as they require labeled OOD data to find the line, but still relevant.

\paragraph{Results}

We summarize the linear correlation results in Table \ref{tab:corr-results}. We can see that COT consistently outperforms all other methods in terms of $R^2$ and $\rho$. Across different datasets and architectures, COT is able to maintain $R^2 > 0.987$ and $\rho > 0.992$. From Fig. \ref{fig:correlation}, we observe that COT gives better estimates than ProjNorm when the error is small, and outperforms Entropy when the error is large. In general, we find that when the shifts are large, COT provides more consistent estimates than AC, Entropy, GDE, and ATC, which often underestimate the error. 

As for the estimation error measured by mean absolute error (MAE), we compare COT with AC, GDE, ATC, which provide direct error estimates. We summarize the results in Table \ref{tab:mae-results} and Fig. \ref{fig:mae}. Again, we see that COT demonstrates notable improvements over previous methods, especially when encountering large shifts. In addition, we would like to point out two observations. Firstly, while COT performs consistent estimates over different architectures, other methods often have difficulty estimating VGG11's error despite similar source performance. This makes COT more broadly applicable to different architectures. Meanwhile, it would be interesting to further understand the root of such discrepancy. Secondly, in our experiment setting, GDE seems to systematically underestimate the test error. The reason is likely that both models started from the same pretrained weights, so the only stochasticity we could impose was the training data order. One would expect the two models fine-tuned from the same pretrained weights might be more similar to each other than two with different initialization trained from scratch are. We plan to perform experiments with models trained from scratch in the future, but we do want to point out that this phenomenon undermines the practicality of GDE as transfer learning is frequently used to improve models' performance. 

\begin{figure}[]
    \centering
    \includegraphics[width=0.9\linewidth]{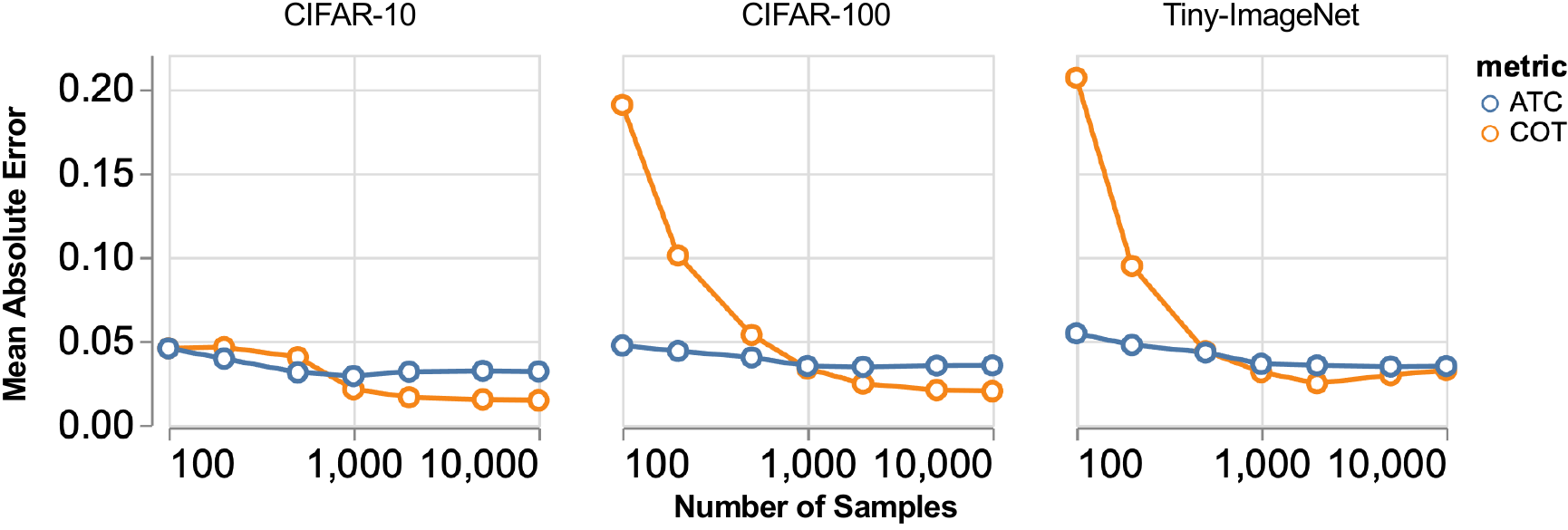}
    \caption{We analyzed the sample complexity of COT and compared with ATC. We found that COT is more sensitive to the number of test samples, but outperforms ATC with 1000 or more samples.}
    \label{fig:sc}
\end{figure}

\paragraph{Sample Complexity} \label{para:sc} In this section, we answer the question: how many test samples are required for our method to provide accurate error estimates? We vary the number of test samples from 100 to 10000 and conduct the experiments using the ResNet50 architecture. The main results are summarized in Fig \ref{fig:sc}. We can see that COT can still outperform ATC with as few as 1000 samples. However, COT deteriorates when there are only 1 or 2 samples for each class (CIFAR-100 and Tiny-ImageNet). This is expected as an good empirical estimation of distribution distances would require a sufficient sample size. Empirically, COT works the best when there are at least 10 samples for each class on average. By contrast, ATC is insensitive to the number of test samples, which is unsurprising, since ATC uses a score function to reduce the softmax outputs to a scalar and adopts a thresholding strategy to estimate the error. To summarize, COT provides superior error estimates when there are at least 2000 test samples. While computing the EMD generally requires a computation time of $\mathcal{O}(n^3)$, current open-source solvers \cite{flamary2021pot} can find the solution of this scale ($n=2000$) within seconds. When we have a large number of samples from the target domain, we can essentially break them down into small batches and take the average of batch estimates as the final estimate.

\section{Conclusion and Future Work} 
In this work, we proposed COT, a novel method providing robust estimations of a model's performance on out-of-distribution target domain without any additional annotation. COT consistently outperforms existing methods on three synthetic shifts datasets across various architectures. For next steps, we plan to evaluate our method on natural shifts due to difference in the image collection process. In terms of computational complexity, while solving for the EMD in COT requires cubic time, we showed that COT only requires a batch size of 2000 to provide superior error predictions. Thus, simply estimating with COT on subsets instead of the whole dataset allows us to scale it to a large number of test samples. In future work, we plan to improve COT's sample complexity to make COT more efficient. Finally, we admit that the current formulation COT is mostly based on heuristics and empirical observations. We are seeking a more formal understanding of why COT works well.

\bibliography{reference}
\bibliographystyle{iclr2023_conference}

% \appendix
% \section{Appendix}

\end{document}